\documentclass[onecolumn]{IEEEtran}
\IEEEoverridecommandlockouts
\usepackage{cite}
\usepackage{todonotes}
\usepackage{setspace}
\usepackage{url}
\usepackage{comment}
\usepackage{todonotes}
\usepackage{amsmath,amssymb,amsfonts}
\usepackage{algorithmic}
\usepackage{graphicx}
\usepackage{textcomp}
\usepackage{xcolor}
\newcommand{\pv}[1]{\textbf{ {\textcolor{magenta}{#1 -- PV}}}}

\usepackage{amssymb}
\def\BibTeX{{\rm B\kern-.05em{\sc i\kern-.025em b}\kern-.08em
    T\kern-.1667em\lower.7ex\hbox{E}\kern-.125emX}}
\begin{document}
\title{Leveraging graph neural networks for supporting Automatic Triage of Patients.
}

\author{\IEEEauthorblockN{Annamaria Defilippo}\\
\IEEEauthorblockA{\textit{Dept. Medical and Surgical Sciences} \\
\textit{Data Analytics Research Center}
\textit{Magna Graecia University of Catanzaro}\\
Catanzaro, Italy \\
 ORCID: 0000-0001-5542-2997\\}
\and
\IEEEauthorblockN{Pierangelo Veltri}\\
\IEEEauthorblockA{\textit{DIMES} Department of Informatics, Modeling, Electronics and Systems \\
\textit{UNICAL}\\
Rende, Cosenza\\
pierangelo.veltri@unical.it}\\
\and
\IEEEauthorblockN{Pietro Li\'o}\\
\IEEEauthorblockA{\textit{Department of Computer Science and Technology } \\
\textit{Cambridge University}\\
Cambridge, UK \\
pl219@cam.ac.uk\\}

\and
\IEEEauthorblockN{Pietro Hiram Guzzi}\\
\IEEEauthorblockA{\textit{Dept. Medical and Surgical Sciences} \\
\textit{Magna Graecia University of Catanzaro}\\
Catanzaro, Italy \\
hguzzi@unicz.it\\}
}

\maketitle

\begin{abstract}
Patient triage plays a crucial role in emergency departments, ensuring timely and appropriate care based on correctly evaluating the emergency grade of patient conditions. Triage methods are generally performed by human operator based on her own experience and information that are gathered from the patient management process. Thus, it is a process that can generate errors in emergency-level associations. Recently,  Traditional triage methods heavily rely on human decisions, which can be subjective and prone to errors. Recently, a growing interest has been focused on leveraging artificial intelligence (AI) to develop algorithms able to maximize information gathering and minimize errors in patient triage processing.  We define and implement an AI-based module to manage patients' emergency code assignments in emergency departments. It uses 
emergency department historical data to train the medical decision process. Data containing relevant patient information, such as vital signs, symptoms, and medical history, are used to accurately classify patients into triage categories. Experimental results demonstrate that the proposed algorithm achieved high accuracy outperforming traditional triage methods. 
By using the proposed method we claim that healthcare professionals can
predict severity index to guide patient management processing and resource allocation.


\end{abstract}

\begin{IEEEkeywords}
component, formatting, style, styling, insert
\end{IEEEkeywords}

\section*{Visual Abstract}

\begin{figure}[ht]
    \centering
\includegraphics[width=0.9\textwidth]{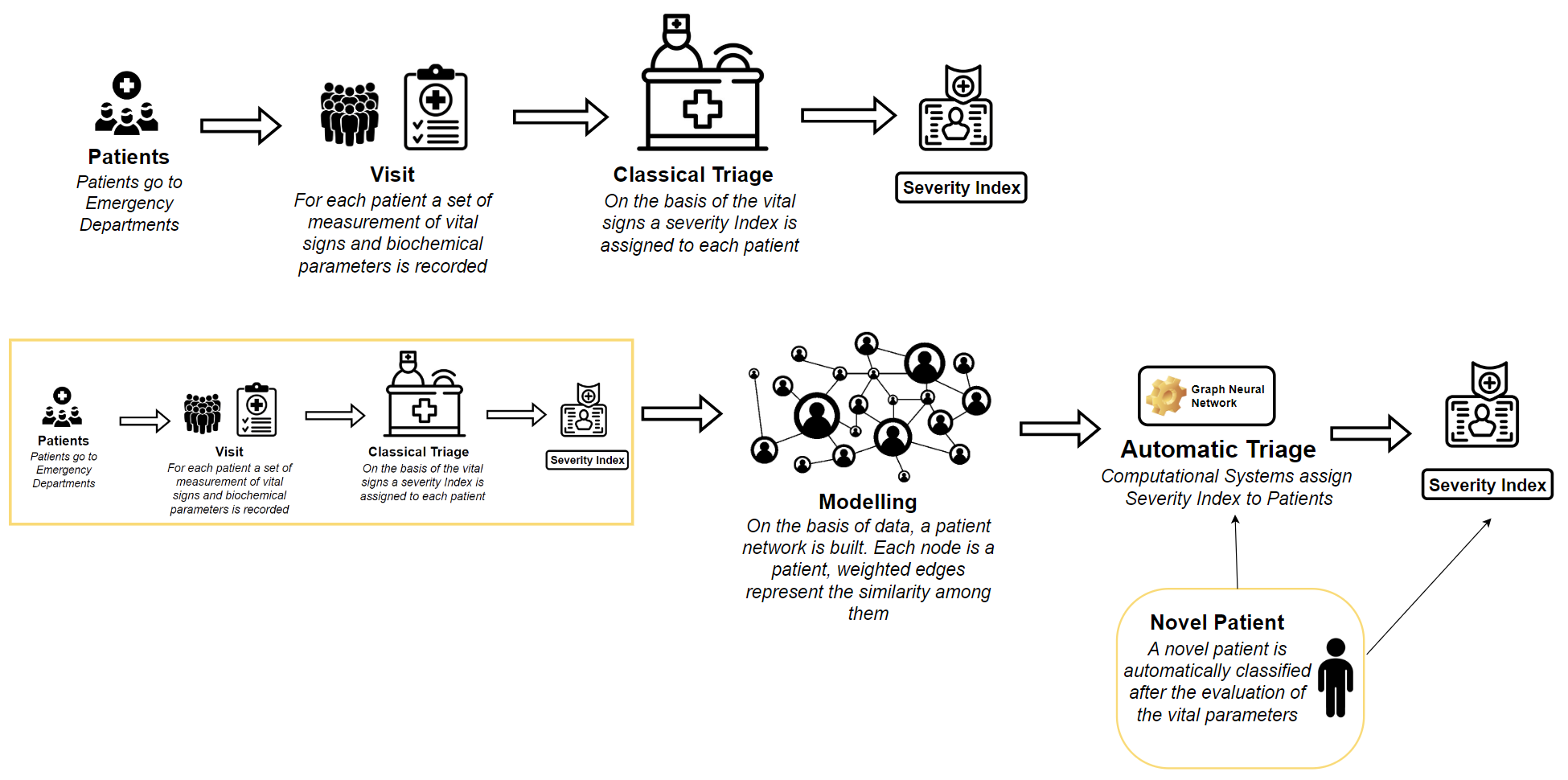}
    \caption{Figure compares automatic vs traditional triage systems. In the traditional triage system, patients go to the Emergency Departments for admission. A set of vital signs and biochemical parameters are evaluated and an emergency code is assigned. In the scenario, we envision patient data from previous admissions are used initially to build a network representing similarity among patients. The network is then used to learn a Graph Neural Network to  classify patients in a latent space (the assignment of the severity index is translated into a multiclass classification problem). Finally, each patient who enters the emergency room is automatically classified after the acquisition of clinical data.} %
    \label{fig:patientflow}
\end{figure}

\section{Introduction}
\label{sec:introduction}

Emergency department (ED) management faces a significant challenge in managing the influx of people. Properly managing queues can help improve the quality of hospitals and contain costs and reimbursement \cite{moschetti2018health}. The management of the patient queues in the ED follows the rules defined by national laws of the healthcare system. In Italy, healthcare services are accessible to all citizens, 
without any restriction. In parallel, in recent years, due to demographic modifications, and pandemics, emergency departments are often overcrowded, even in the absence of exceptional conditions such as pandemics, highlighting the relevance of the problem \cite{combi2023ihi,parmeggiani2010healthcare,savioli2022emergency,bambi2021new}. In general, managing queues to emergency departments (EDs) is crucial, considering both economic and social aspects due to the potentially fatal outcomes resulting from the unfair prioritisation of patients. The prioritisation in the patient management process regarding cure and study is based on a process called \text{triage}. The triage consists of gathering information from a first rapid patient visit, measuring vital signs and parameters, thus defining a grade of emergency for each patient and indicating a class of treatment priority. Each priority class is assigned to an emergency code associated with a patient priority code and a related treatment protocol.
Therefore, correct prioritisation of queues for new patients (i.e., classification) is based on the clinical and health status conditions of patients and should be able to guarantee the proper prioritisation treatment to patients incoming from the emergency department \cite{cremonesi2015robustness,boissin2022clinical}.

Methods and techniques for triage definition have been defined and adopted at a large scale level  \cite{defilippo2023computational,hinson2019triage}. The Italian rules of the triage process rules have been defined in  \textit{the Italian healthcare management guidelines}, where triage is composed of  four phases: 
\begin{enumerate}
    \item \textit{Immediate Evaluation Phase} (so-called on the door): the identification of the conditions of the patients to screen  people who need immediate intervention;
    \item   \textit{Phase of subjective and objective evaluation}: evaluation of condition through anamnestic interviews and detection of clinical signs and vital parameters through clinical analysis.
    \item \textit{Triage Decision Phase}: assignation of a priority code for each patient and implementation of waiting queues;
    \item \textit{Re-evaluation Phase}: the periodic analysis of patients to confirm or modify the priority code.
\end{enumerate}

They were taking into account the worldwide scenario \cite{meyer2013physicians,bambi2021new}, some of the most adopted systems are the Canadian Triage and Acuity Scale (CTAS) \cite{j2003canadian,bullard2017revisions} the Australasian Triage System (ATS) \cite{putri2020australasian}, the Manchester Triage System (MTS) \cite{azeredo2015efficacy}, the Emergency Severity Index (ESI, United States) \cite{wuerz2001implementation}, the Korean Triage and Acuity Scale (KTAS) \cite{kwon2019korean}, the Taiwan Triage Acuity Scale (TTAS) \cite{ng2011validation} and the South African Acuity Scale (SAAS) \cite{meyer2018validity}. All such systems share the  \textit{streaming} of patients, which is based on grouping patients using conditions and the subsequent assignment of patients into separate areas of the ED  \cite{meyer2013physicians}.

The application of triage rules is mostly based on operator decisions (both nursery and physician) interacting with patient parameters and data. 
The triage assessment is related to various factors determining the urgency of each incoming patient. 

All the triage systems, based on manual interventions, share some common limitations also investigated by Fitzgerald et al., \cite{fitzgerald2010emergency} which emphasise the inconsistency in triage assessment due to various factors determining the urgency of each incoming patient. {Thus, many approaches have been to defining and introducing computational-based triage code automatic assignment mechanisms to minimise human-related error risks \cite{eid2021comparing,chong2016development}.}

Consequently, many authors explored the possibility of introducing automated systems based on computational intelligence to improve triage systems by suggesting the emergency code \cite{eid2021comparing,chong2016development}.
{Moreover, machine learning (ML) and artificial intelligence (AI) algorithms have been used to analyse and learn from clinical data such as Electronic Medical Records (EMRs) as well as from additional information gathered, for instance, from sensors or wearable devices.} More recently, machine learning (ML) and artificial intelligence (AI) algorithms have shown the possibility of analysing electronic medical records (EMRs) \cite{cheung2019machine,canino2015analysis} as well as unstructured patient data such as sensors and wearable devices. Thus, since triage systems are based on the observation of patient data, the possibility of using ML and AI has been explored to improve classical clinical scoring systems \cite{hinson2019triage}. 

Many prediction models have been developed to enhance the triage process, suggesting a more-grained stratification of patients inside the classical groups and better performances in guaranteeing better clinical outcomes  \cite{defilippo2023computational,hinson2019triage}.
A range of machine-learning approaches have been explored for automatic patient triage in emergency departments. 
Olivia et al., \cite{olivia2018machine} effectively employed various supervised learning algorithms, including Naive Bayes, Support Vector Machine, Decision Tree, and Neural Network, to predict patients' medical conditions accurately. Caicedo-Torres et al ., \cite{caicedo2016machine} expertly utilised machine learning techniques in a  Pediatric ED to determine which patients should be admitted to the Fast Track {(i.e., rapid admission and pediatric treatment)}. Joseph et al., \cite{joseph2020deep} conducted a deep study to identify critically ill patients using deep learning approaches, specifically analysing the use of limited information available at triage. Collectively, these studies demonstrate the immense potential of machine learning in managing patient triage in the emergency department. Existing prediction models are based on data collected at triage, such as demographic information, vital signs, primary complaints, nursing observations, and initial diagnostics  \cite{defilippo2023computational,hinson2019triage}. Historical data such as patient access frequency and medical records frequencies are also used in some predicting models \cite{salman2021review}. Some models also include variables like hospital visit frequency and previous medical history. {Thus, using data gathered from patient data regarding emergency department admission and treatment, it is possible to enrich and improve triage and processes to manage incoming patients.} 
However, data obtained at later stages of a patient's current ED visit, such as laboratory tests, administered medications, and diagnoses made by ED staff, is more effective in predicting admissions. 

Modelling similarities among patients' conditions and similarities among previous visits by joining for the same patient may support the definition of an admission prediction at the triage stage. Available solutions often use broad categories for chronic conditions \cite{levin2018machine,leung2021novel}. Enriching EHR data to train models could improve prediction results \cite{hong2018predicting}. Moreover, tentative based on gradient boosting and deep neural networks improve prediction efficacy. Nevertheless, all the existing approaches need to include the explicit modelling of the similarity among patients \cite{bentejac2021comparative}. 

\begin{figure}
    \centering
\includegraphics[width=0.5\textwidth]{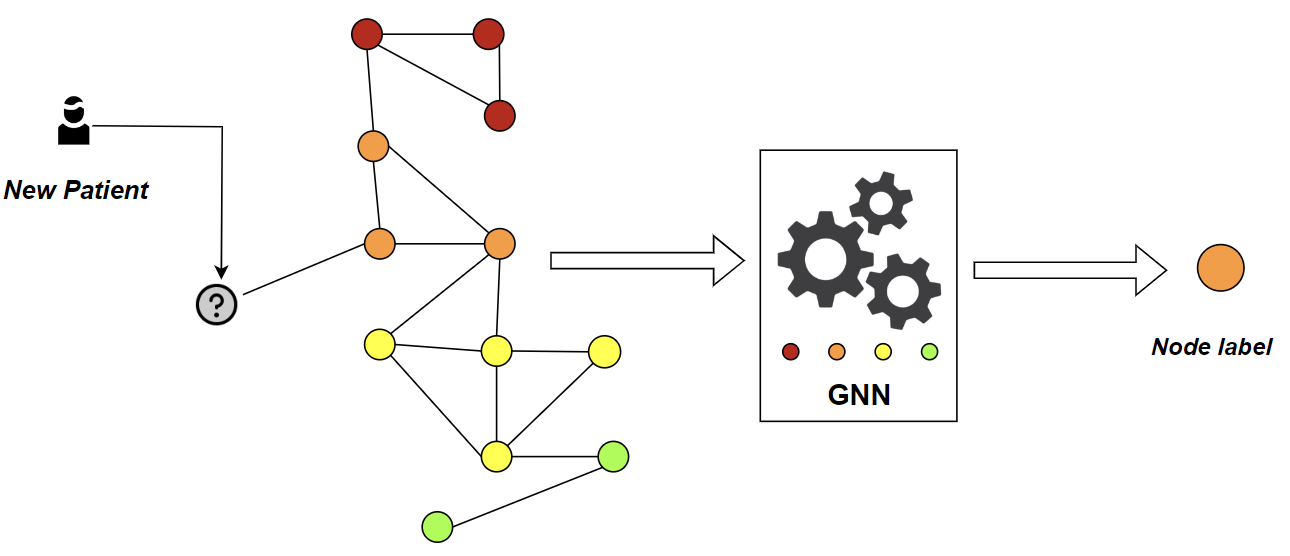}
    \caption{Patient data are used to build a network representing patient similarity. Each node has a previous severity index codified as a node label. The network is then used to learn a Graph Neural Network to classify patients. Therefore, the assignment of the severity index is translated into a multiclass node classification problem. A new patient requiring admission to ED is a node of the graph without a label, and the learned model is used to derive the correct label.}
    \label{fig:shortsummary}
\end{figure}

 We leverage network science methods to build a novel clinical algorithm based on artificial intelligence and network science for assigning priority to patients. 
We start by considering clinical patient data extracted from patient records. These data include analytical observations as well as subjective ones. A modelling phase is applied after an initial preprocessing step to identify possible noise and outliers. Each patient is modelled as a graph node in the learning phase, while edges represent the similarity among the observation data. At this point, the graph is embedded into a latent space. Finally, patients are classified into the risk groups by applying a node classification algorithm as summarised in Figure \ref{fig:shortsummary}.

Prior models incorporating a patient's medical history often use broad categories for chronic conditions. Recently, a recent study \cite{hong2018predicting} demonstrated that using comprehensive EHR data could robustly predict in-patient outcomes. Thus, a predictive model for hospital admission that leverages a thorough compilation of a patient's history could surpass earlier models in effectiveness.

Although gradient boosting and deep neural networks have been recognised as potent tools in predictive modelling, all the existing approaches lack in modelling the similarity among patients \cite{bentejac2021comparative}. Building upon previous research, we leverage network science methods to build a novel clinical algorithm based on artificial intelligence and network science for assigning priority to patients. We start by considering clinical patient data extracted from patient records. These data include analytical observations as well as subjective ones. A modelling phase is applied after an initial preprocessing step to identify possible noise and outliers resulting in the building of a network. Each patient is modelled as a graph node in this step, while edges represent the similarity among the observation data. At this point, the graph is embedded into a latent space. Finally, patients are classified into risk groups by applying a node classification algorithm as summarized in Figure \ref{fig:shortsummary}.

 Since the embedding is inductive, when a novel patient is available, a novel node is added to the graph and embedded \cite{zitnik2023current,guzzi2012semantic}. We tested our pipeline on public data to demonstrate the effectiveness of our approach and the improvement concerning state-of-the-art approaches. 

\subsection{A Toy Example}
\label{sec:toyexample}

This section describes the system through a toy example. In a classical triage system, patients requiring admission to ED are classified into a severity level using standard algorithms which produce a severity index based on the patient data. Such data may be given by the observation of the patient by the nurse, and evaluation of biochemical parameters. Such data are usually stored in hospital registries and EHRs. We start by considering these data to build a patient similarity network where each patient is a node. The edges among patients are weighted by calculating the similarity of patients on the basis of their data. Each node is also labeled with the previously assigned severity level. The network is embedded into a latent space and a node classifier is trained. Once that the classifier is trained, each new patient can be automatically assigned to a severity index by using the classifier.

\section{Related Work}
\label{sec:related}

Triage is a critical and systematic process used in an emergency department (ED). It helps prioritize patients based on the severity of their condition, the urgency of their need for care, and the availability of resources within the healthcare facility. The goal is to ensure that patients who need immediate attention receive it promptly, while those with less critical needs are attended to in an order that maximizes the overall efficiency and effectiveness of emergency medical services.

The triage process in an emergency department involves several key steps and principles, which are as follows.

Upon arrival at the ED, each patient undergoes an initial assessment by a triage nurse or a trained healthcare professional. This assessment is designed to quickly gather critical information about the patient's condition, including the chief complaint, vital signs (such as temperature, blood pressure, heart rate, and respiratory rate), and a brief history of the present illness or injury.

Urgency Categorization: Based on the initial assessment, patients are categorized into different levels of urgency ranging from immediate life-saving intervention needed (highest priority) to non-urgent care (lowest priority).

 Triage categorization directly influences the allocation of resources. Patients with the highest urgency levels are treated immediately and often directed to specialized areas within the ED equipped to handle severe cases (e.g., resuscitation rooms). Those with lower urgency levels may wait longer and be seen in order of priority based on their triage category.

 Triage is not a one-time assessment but a continuous process. Patients' conditions can change, necessitating a reassessment and potential re-categorization of their urgency level. Triage nurses or other designated healthcare professionals periodically re-evaluate waiting patients to ensure that changes in their condition are promptly identified and addressed.

Accurate documentation and effective communication are essential components of the triage process. Triage decisions, patient information, and any changes in condition must be thoroughly documented and communicated to the entire healthcare team involved in the patient's care. This ensures continuity of care and that all team members are aware of the patient's status and needs.

 Triage involves ethical considerations, such as fairness, equity, and the principle of doing the most good for the greatest number of people. Healthcare professionals must make unbiased decisions based on clinical urgency and the potential for benefit from medical intervention, rather than factors like financial status, age, or social position.

 The triage process faces various challenges, including overcrowding in emergency departments, fluctuating patient volumes, and limited resources. Effective triage requires flexibility and the ability to adapt to changing circumstances, such as public health emergencies or disasters, which may necessitate modifications to triage protocols and prioritization strategies.


Despite the worldwide adoption and diffusion, some common problems affect the triage systems:  such as dependence on subjective medical staff assessment and the possibility of having many missing variables. 


Conversely in an automatic triage system, the categorization of patients is fully automated and the level of severity is determined by computer algorithms  as summarised in  \cite{levin2018machine}.

Automatic (or computating based),  triage systems present some advantages such as: (i) stability of assignment \cite{levin2018machine}, (ii) filtering out noise in the patient variables, (iii) model and analyse patient similarity,  (i.e. by modelling the set of patients in a network which evidences patient similarity \cite{yu2020machine,levin2018machine,choi2019machine,cannataro2010impreco,hiram2022disease}; (iv) avoiding  patients under triaged into low severity levels \cite{inokuchi2022machine}; (vi) avoid  of racial, gender, age bias \cite{allen2020racially}. 
Literature contains many approaches of the use of machine learning for patients emergency classification.  Singh et la., \cite{singh2018machine} developed a cascading classifier for psychiatric patient triage, achieving high accuracy and reducing expert effort. Graca (2023) highlighted the potential of machine learning in ICU triage and patient transfer during crises, such as the Covid-19 pandemic. Olivia et al., \cite{olivia2018machine} and Yan et al., \cite{yan2019technology} both emphasized the significance of machine learning in emergency department triage, focusing the effectiveness of Support Vector Machine and Decision Tree models. 

\section{Architecture}
\label{sec:arch}

\begin{figure}[ht]
    \centering
    \includegraphics[width=0.7\textwidth]{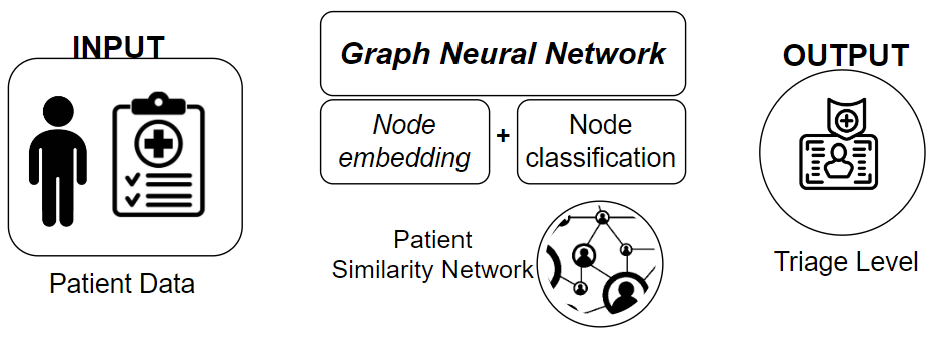}
    \caption{The Architecture of the System.}
    \label{fig:architecture}
\end{figure}

Figure \ref{fig:architecture} describes the architecture of the system.
The system receives as input patient data to build the patient similarity network where each node is a patient and the weighted edges model the similarity among them. The current implementation of the system uses four known measures for evaluating the similarity: cosine similarity, Euclidean,  Manhattan, and Minkowsky distances. 

The network embedding (or graph representation learning) module, is responsible for projecting the network into a latent space by using node embedding. GRL projects each node into a separate point in a subspace while preserving the initial distance between nodes.

The node classification module is responsible for the assignment of a triage level to each new patient requiring the assignment. Both the graph representation learning  module and node classification module leverage the computational intelligence of Graph Neural Networks methods.  From the existing methods for GRL we select those which are able to deal with node labels and based on inductive strategies. For instance, Graph Neural Networks are used for node embedding \cite{guzzi2022editorial} because the present two main advantages:  (i) they  take into account data related to nodes, (i.e. node features); (ii) they are inherently inductive, so they do not need to recalculate the whole embedding in case of any graph modification (e.g. node/edge insertion or removal) \cite{gu2022modeling,guzzi2012semantic,kumar2021data}. 
The current implementation uses three state of the art methods:Graph Convolutional Networks \cite{betkier2023pocketfindergnn}, GATv2Conv \cite{brody2021attentive}, and GraphSage \cite{hamilton2017inductive,guzzi2023analysis}, which in general have better performances compared to the state of the art.

Graph Convolutional Networks (GCNs) \cite{zhang2019advanced} are specialized neural network architectures for processing data structured as graphs. These networks are particularly effective in scenarios where data points are interconnected, such as social networks, molecular structures, or communication networks. Unlike traditional convolutional neural networks that operate on grid-like data (e.g., images), GCNs leverage the graph structure to process data on nodes and their connections \cite{zhang2019advanced}. They apply convolution operations directly on the graph, aggregating information from a node's neighbours capturing both the features of individual nodes and the complex relationships between them. The flexibility of GCNs in handling irregular graph structures makes them highly suitable for tasks like node classification, link prediction, and graph classification. 


Graph Attention Networks (GATs) \cite{velickovic2017graph} use an attention mechanism to dynamically determine the relevance of each neighbor's features, allowing for a more nuanced and context-aware aggregation of neighbor information. This approach is different from traditional graph convolutional networks that aggregate neighboring node features in a uniform or predetermined manner. 

{{In GATs, the attention mechanism assigns different weights to more relevant nodes in a neighborhood, which is particularly useful in dealing with complex graph structures where the relevance of neighboring nodes can vary significantly. This approach leads to more effective feature representation and learning. 
The implementation of GAT is defined as followed with two GATv2Conv layers:
\begin{itemize}
\item  first layer with input dimension equal to 16 as the number of features, output dimension equal to 8 and 4 attention heads to implement the mechanism of attention. 
\item  second layer with input dimension equal to the output dimension of the previous layer (hidden dimension * number of attention heads) to obtain the output of dimension 4 as the size of the target class.
\end{itemize}
}}


GraphSAGE \cite{hamilton2017inductive} is a framework for generating node embeddings for large graphs. Unlike conventional graph-based methods, it samples and aggregates from a node's local neighborhood to efficiently scale to large graphs. It can perform aggregation using various functions and capture diverse neighborhood structures effectively. GraphSAGE has inductive learning capability and is effective in classification, prediction, and recommendation systems for large-scale and dynamic graph data.


{}

\section{Experimental Results}

\begin{figure}
    \centering
    \includegraphics[width=0.75\textwidth]{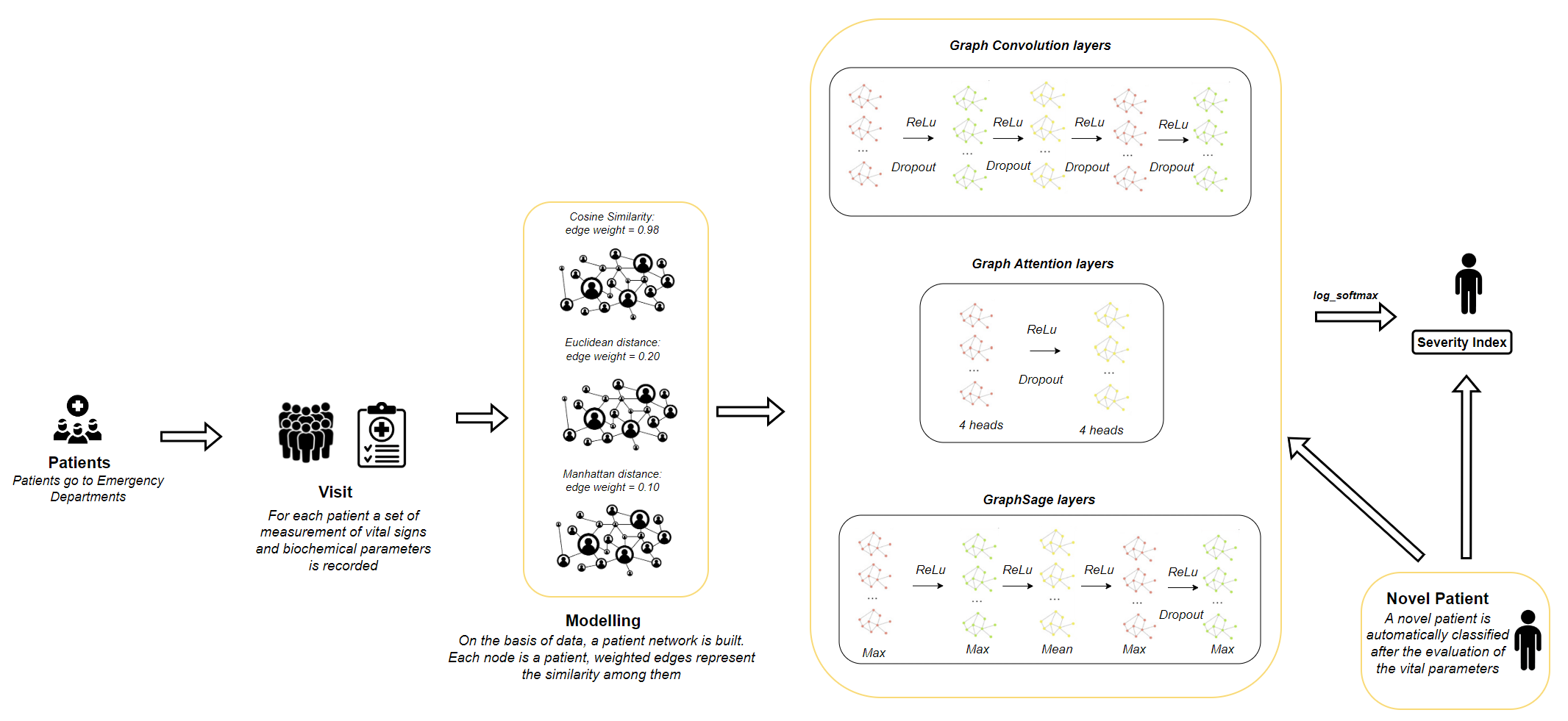}
    \caption{The Figure represents the experiments we performed to test the approach. The input dataset is converted into a network using cosine similarity, Euclidean, Manhattan, and Minkowski Distances. For each measure, we generated a set of networks using different thresholds. Then each single network is used to learn a classifier and finally results are evaluated.
    }
    \label{fig:experimentalflow}
\end{figure}

To test the performances of our method we designed and performed a set of experiments as depicted in Figure \ref{fig:experimentalflow}. The input dataset is translated into networks using cosine similarity, Minkowski, Manhattan, and euclidean distance.  We built  many  networks for each measure using different threshold levels. Then for each network  we perform node embedding and we train a a classifier using GCN, GAT and GraphSage. Finally, we evaluated the performance of each classifier.

\subsection{Dataset}
 We tested our methods on a publicly available dataset  on the Kaggle platform (\url{https://www.kaggle.com/datasets/hossamahmedaly/patient-priority-classification}). The dataset contains 6962 instances (rows) of patient admission  and 16 features for each instance.  Each row  describes the parameters of a patient described  such as symptoms and some biochemical parameters used to determine the severity level as summarised in Table \ref{tab:features}. 
 Each row also contains the assigned the triage level as follows with decrescent level of risk:
\begin{itemize}
\item \textit{Red}: the patient needs immediate attention;
\item \textit{Orange}: the patient needs intervention in a short time;
\item \textit{Yellow}: urgent condition needing interventions that  can be deferred;
\item \textit{Green}: condition with minor urgency because there are no alterations of vital functions and no critical symptoms. 
\end{itemize}

\begin{table}[h]
    \centering
     \caption{Features of the patients.}
  \begin{tabular}{|l|l|}
\hline
\textbf{Parameter} & \textbf{Description} \\
\hline
Age & age of the patient. \\
\hline
Gender & patient's sex. \\
\hline
Chest pain type & the type of chest pain. \\
\hline
Blood pressure & blood pressure value. \\
\hline
Cholesterol & cholesterol level. \\
\hline
Max heart rate & maximum heart rate value. \\
\hline
Exercise angina & presence of angina. \\
\hline
Plasma glucose & glucose level in blood plasma. \\
\hline
Skin thickness & any thickening of the skin. \\
\hline
Insulin & insulin level. \\
\hline
BMI & body mass index \\
\hline
Diabetes Pedigree & genetic predisposition to diabetes. \\
\hline
Hypertension & elevated blood pressure. \\
\hline
Heart disease & presence of heart disease. \\
\hline
Residence type & type of residence place. \\
\hline
Smoking status & defines whether or not the patient is a smoker. \\
\hline
Residence type: & Urban,Rural \\ \hline
Smoking status:&  never smoked,smoke, previously smoked,Unknown.\\ \hline
\end{tabular}
   
    \label{tab:features}
\end{table}


Data are pre-processed as follows:

\begin{enumerate}
    \item \emph{Duplicates and null values}: null values are removed from the dataset, reducing the number of entries in the patient tables. 
    \item \emph{Inconsistent (or incomplete) records}: missing values are replaced with the mode of the values of the whole corresponding column. 
        \item \emph{Label Encoder for categorical features}: Dataset contains three categorical features: \textit{Residence type}, \textit{Smoking status} and  the third is the target feature. 
    \item \emph{Oversampling and Undersampling}: Since classes are unbalanced, we use  SMOTE (Synthetic Minority Oversampling Technique)  \cite{chawla2002smote} to perform the sampling. 
    \item \emph{Normalization with Min-Max Scaler}: Finally data are normalized using a Min-Max approach, so a maximum value of each column is equal to 1 and a minimum value equal to 0. 
\end{enumerate}
\begin{figure*} [h]
    \centering
\includegraphics[width=0.7\textwidth,keepaspectratio]{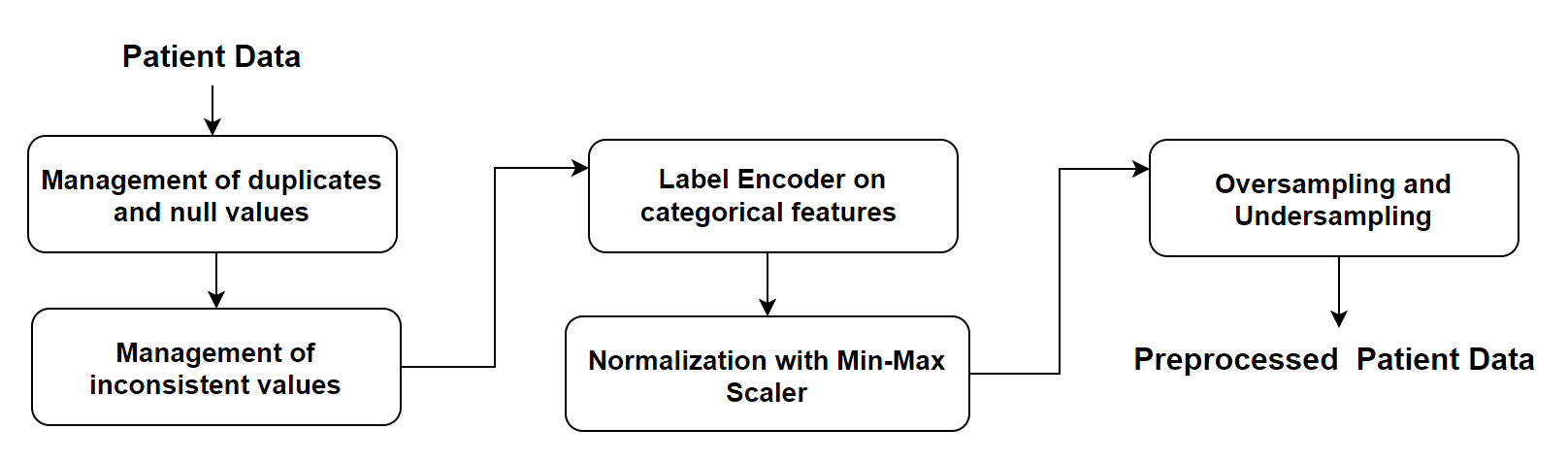}
    \caption{Figure reports the Preprocessing workflow. }
    \label{fig:flow}
\end{figure*}

\subsection{Networks generated using Cosine Similarity}

We first generated patient networks using cosine similarity to define edges. We used different thresholds (0.98, 0.95, 0.94,0.92,0.90). In this case an edge connect two nodes when the cosine similarity is greater than the threshold. We stopped at 0.90 since we reached a completely connected graph.
Table \ref{tab:cosineNet2} summarises the characteristics of the networks generated by using cosine similarity at different level of threshold.

\begin{table}[h]
    \centering
      \caption{Characteristics of the networks generated by using different levels of thresholds and cosine similarity as measure.}
    \begin{tabular}{|c|c|c|}\hline
      Threshold  & Isolated Nodes & Edges  \\
         \hline
         \textbf{0.98} & \textbf{761} & \textbf{1.578.490}\\
         \hline
         \textbf{0.95} & \textbf{22} & \textbf{8.103.196}\\
         \hline
         0.94 & 7 & 10.810.687 \\
         \hline
         0.92 & 2 & 16.505.521 \\
         \hline
         0.90 & 1 & 22.148.695 \\
          \hline
    \end{tabular}
  
    \label{tab:cosineNet2}
\end{table}
As it can be noticed, decreasing the value of the thresholds reduces the isolated nodes and increases the number of edges. In this way, it is possible to connect more patients due to the possibility to create ad an edge between two nodes with some more dissimilarities. 

\subsection{Networks generated using Euclidean Distance}

We first generated patient networks using Euclidean Distance to define edges. We used different thresholds (0.20, 0.23, 0.25,0.28,0.31, 0.38). In this case an edge connect two nodes when the euclidean distance is lower than the threshold. We stopped at 0.38 since we reached a completely connected graph.

Table \ref{tab:euNet} summarises the characteristics of the networks generated by using Euclidean Distance at different level of threshold.
\begin{table}[h]
    \centering
       \caption{Characteristics of the networks generated by using different levels of thresholds and Euclidean Distance as similarity measure.}
     \begin{tabular}{|c|c|c|}\hline
      Threshold  & Isolated Nodes & Edges  \\
      \hline
         \textbf{0.20} & \textbf{125} & \textbf{3.348.231}\\ 
       \hline
         \textbf{0.23} & \textbf{33} & \textbf{5.403.317} \\ 
        \hline
         0.25 & 14 & 7.120.567\\
         \hline
         0.28 & 5 & 10.279.280\\
         \hline
         0.31 & 0 & 14.086.639\\
         \hline
         0.38 & 0 & 25.226.436\\
         \hline
    \end{tabular}
 
    \label{tab:euNet}
\end{table}
The considerations are similar to the case concerning cosine similarity, with the difference that the number of isolated nodes is is lower than the first obtained previously. It may be associated with the fact that considering distance allows for a greater number of connections between a larger number of different nodes.

\subsection{Networks generated using Manhattan Distance}

Table \ref{tab:manhNet} summarises the characteristics of the networks generated by using Manhattan Distance at different level of threshold.
\begin{table}[h]
    \centering
    \caption{Characteristics of the networks generated by using different levels of thresholds and Manhattan Distance as  similarity measure.}
     \begin{tabular}{|c|c|c|}\hline
      Threshold  & Isolated Nodes & Edges \\
      \hline
         \textbf{0.10} & \textbf{947} & \textbf{1.186.583}\\
         \hline
         \textbf{0.13} & \textbf{175} & \textbf{2.838.619}\\
         \hline
         0.22 & 0 & 14.688.023\\
         \hline
         0.31 & 0 & 35.306.331\\
         \hline
         0.33 & 0 & 41.217.110\\
         \hline
    \end{tabular}
    
    \label{tab:manhNet}
\end{table}
In this scenario, initially, the number of edges is lower than in previous cases, as the number of isolated nodes. On the contrary, after the first two, much less stringent thresholds have been chosen causing a substantial increase in the number of edges and a reduction to zero in the number of isolated nodes.

\subsection{Networks generated using Minkowsky Distance}
In this subsequent phase, it is interesting to generate graphs from tabular data, by exploring the Minkowski distance. It serves as a generalization of both Euclidean and Manhattan distances, therefore, it is appropriate to vary the value of p, which characterizes the modification of this metric.  

Table \ref{tab:minkNet10} summarises the characteristics of the network generated by using Minkowsky Distance (p = 10) at different level of threshold.
\begin{table}[h]
    \centering
     \caption{Characteristics of the network generated by using different levels of thresholds and Minkowsky Distance (p = 10) as similarity measure.}
     \begin{tabular}{|c|c|c|}\hline
      Threshold  & Isolated Nodes & Edges  \\
      \hline
         \textbf{0.20} & \textbf{892} & \textbf{988.835}\\
         \hline
         \textbf{0.25} & \textbf{225} & \textbf{2.146.676}\\
         \hline
         0.30 & 83 & 3.942.436 \\
         \hline
         0.35 & 37 & 6.403.802\\
         \hline
         0.40 & 13 & 9.481.995\\
         \hline
    \end{tabular}
   
    \label{tab:minkNet10}
\end{table}

Table \ref{tab:minkNet4} summarises the characteristics of the network generated by using Minkowsky Distance (p = 4) at different level of threshold.
\begin{table}[h]
    \centering
      \caption{Characteristics of the network generated by using different levels of thresholds and Minkowsky Distance (p = 4) as  similarity measure.}
     \begin{tabular}{|c|c|c|}\hline
      Threshold  & Isolated Nodes & Edges  \\
      \hline
         \textbf{0.20} & \textbf{394} & \textbf{1.630.369}\\
         \hline
         \textbf{0.25} & \textbf{92} & \textbf{3.527.312} \\ 
         \hline
    \end{tabular}
  
    \label{tab:minkNet4}
\end{table}

It may be observed that choosing p = 10 results in a notable reduction in network connections, differently from choosing p = 4 in which the values are closer to other measures. 

\subsection{Graph Convolutional Networks}
{{We used GCN to  analyze graphs created based on Cosine similarity, and Manhattan on Euclidean distance. We employed two architectures of GCN, the first one for cosine and Manhattan and the last one for the networks generated using Euclidean distance.

The first architecture is composed by five GCNConv layers: 
\begin{itemize}
\item layer 1 of dimensions (16,64), where the first one represents the input feature dimension for each node in the graph: each node is described by a vector of 16 features.
\item layers 2,3,4 of dimensions (64,64) representing hidden layers’ dimension during GCN convolutions, followed by dropout function with a fraction of characteristics to be reset equal to 20\%, during training and ReLu function for each layer. 
\item layer 5 of dimensions (64,4) with an output dimension appropriate to the number of classes to be identified as targets.
\end{itemize}
Similarly, the second architecture is composed in the same way, excluding the fifth layer and using 32 as the hidden dimension during GCN convolutions. The input size is set to 16 (in agreement with the number of features), while the output equal to 4 (in agreement with the number of feature target's classes). Furthermore, the same number of Dropout layers are inserted with the same fraction of characteristics to be reset.
}}

\subsection{Graph Attention Networks}
The implementation of GAT is defined as followed with two GATv2Conv layers:
\begin{itemize}
\item  first layer with input dimension equal to 16 as the number of features, output dimension equal to 8 and 4 attention heads to implement the mechanism of attention. 
\item  second layer with input dimension equal to the output dimension of the previous layer (hidden dimension * number of attention heads) to obtain the output of dimension 4 as the size of the target class.
\end{itemize}
The two Dropout layers have been introduced with a rate of 20\% to improve the performances and to reduce the overfitting.

\subsection{GraphSage}
{{Firstly, the available data is further subjected to a phase which consists in the creation of batches to be able to apply the training on the mini-batches that are better manageable for training the GNN. The  \textit{torch geometric.loader} module provides the \textit{Neighborloader} function with the possibility to choose both the size of each batch and the number of neighbours to consider at each iteration. Five sub-graphs are obtained for each considered graph choosing a dimension equal to 3000 for the batch size and a number of neighbours equal to 10 to be considered at each iteration for each node for 5 iterations.
For all the graphs created by the similarity and distance measures, the same model is implemented using five SAGEConv layers. SAGEConv is better than simple GraphSage allowing to capture further details regarding the graph structure, thanks to its aggregate representation based on the degree of the nodes in the neighborhood. In the current implementation, each layer has an input dimension equal to the output dimension of the previous layer, with a sequence of 64, 32, 16, 8. Naturally, the input dimension of the first layer always reflects the number of features, and the output dimension of the last layer corresponds to the number of possible values for the target. Additionally, each SAGEConv layer uses the max pooling aggregation function, except for the third layer with mean aggregation function. 
Similar to the architectures described earlier, each layer applies a ReLU activation function after the aggregation. Also, a dropout of 20\% is applied, but only after the fourth layer. }}
\subsection{Classification Performances}

\begin{figure*}[h]
    \centering
 \includegraphics[width=0.55\textwidth,keepaspectratio]{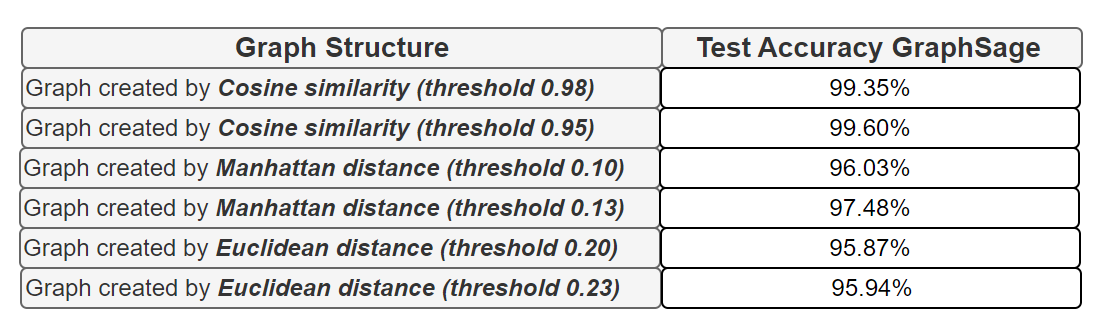}
    \caption{Comparison of test accuracy considering an additional threshold for each metric.}
    \label{fig:comp1}
\end{figure*}

\begin{figure*}[h]
    \centering
 \includegraphics[width=0.55\textwidth,keepaspectratio]{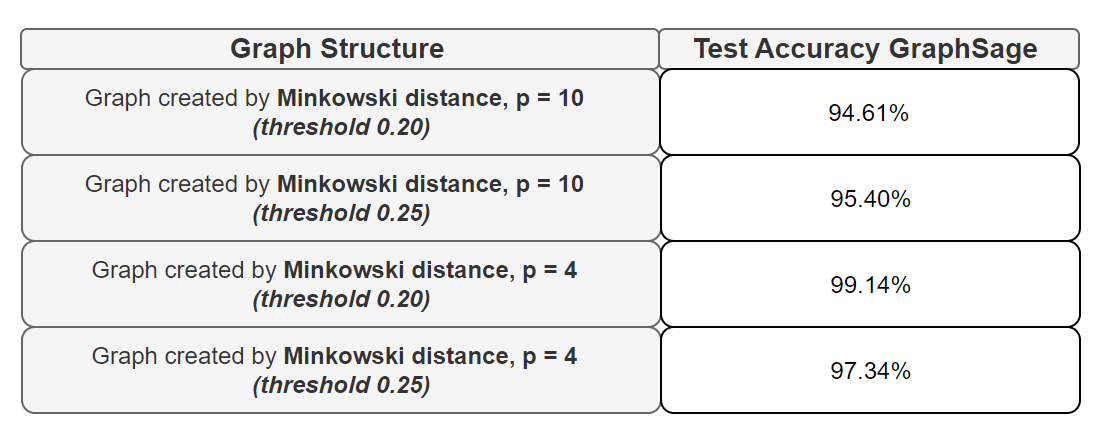}
    \caption{Comparison of test accuracy considering an additional metric.}
    \label{fig:comp2}
\end{figure*}

We  measured the classification performances for each network previously built. This helped us to study the impact of how some variations in parameters of the system affect the conclusions of a model.  Indeed, the obtained results are visualized in Figures \ref{fig:comp1} and \ref{fig:comp2}. 

Additionally, it can be observed that for all the considered metrics, opting for a less strict threshold enhances the model's performance. The only exception is made for the Minkowski distance with p=4, for which the performance slightly decreases.
For this reason, further experiments were conducted with additional arbitrary thresholds, focusing only on the Minkowski distance with p = 10 and excluding p = 4.

\begin{figure*}[h]
    \centering
 \includegraphics[width=0.80\textwidth,keepaspectratio]{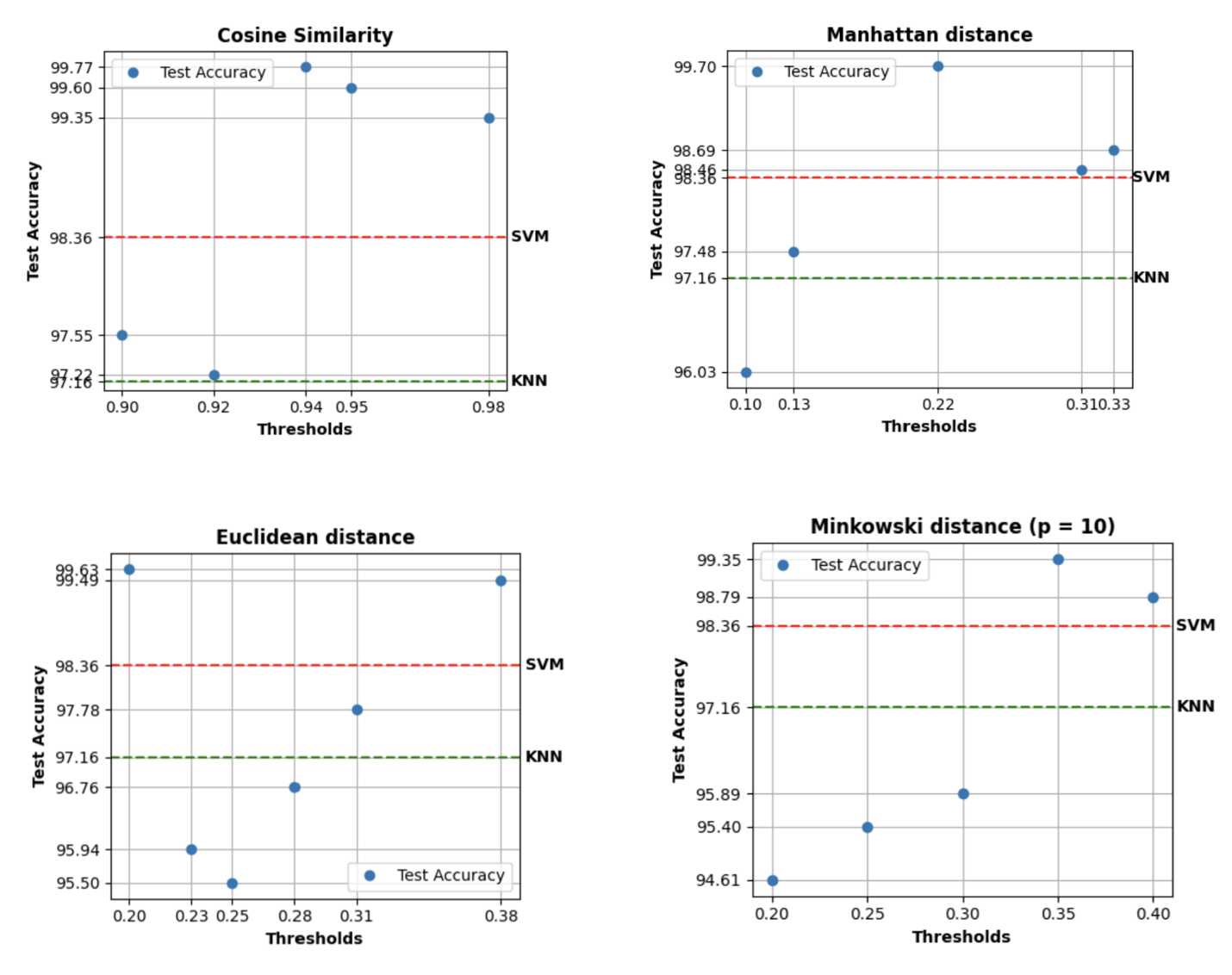}
    \caption{Comparison of test accuracy in relation to the threshold value used for edge creation in each network for each considered metric.}
    \label{fig:secondcomp}
\end{figure*}

Based on the preliminary results obtained, it seems appropriate to increase the number of threshold points, used to create the graph structures shown in the previous tables for each metric respectively. The aim will be to display the performance growth or the performance loss. All the values of the arbitrarily considered thresholds are reported in the previous tables, while in the following section, the resulting performances are presented as illustrated in the figure \ref{fig:secondcomp}.
For each experiment considering a different metric, there are some thresholds that influenced negatively the performances of the models. In fact, these models underperform compared to the others. In addition, the results are worse also than the baseline. On the contrary:
\begin{itemize}
\item For the cosine similarity, choosing a threshold lower by a few points, is positive for the performances of the model, maybe because it helps to find more similarity useful for the classification. On the other hand, if the threshold is lower than 0.94, it isn't good. 
\item Differently from the previous case, an intermediate value of Manhattan distance could be the better choice outperforming also the baseline.
\item Considering the Euclidean distance, higher thresholds deteriorate the results in some cases, as it can be seen, in a more evident manner, for thresholds like 0.23 and 0.25. 
\item Ultimately, for the Minkowski distance, to outperform the baseline, the model requires thresholds to be increased in relation to the initial values.
\end{itemize}

\subsection{Node Classification}

For all the previous methods, we used following parameters to build the classifier:

\begin{itemize}
\item the CrossEntropyLoss that is computed between the model predictions and the training labels to measure how well the model is learning.
\item the Adam optimizer that is employed to update the model weights with a weight decay = 5e-4 and a learning rate of 0.01. This last one is used in all cases, except for the model based on GAT layers, where it has been set to 0.005. 
\end{itemize}

\subsection{Comparison with respect classification on tabular data}

It could be useful to compare the results obtained with networks related to  classification on tabular data. Therefore, to do that, it was trained two types of algorithms: Support Vector Machine and K-Nearest Neighbors on the same patient data used for classification on graphs.
\begin{figure*}[h]
    \centering
 \includegraphics[width=0.50\textwidth,keepaspectratio]{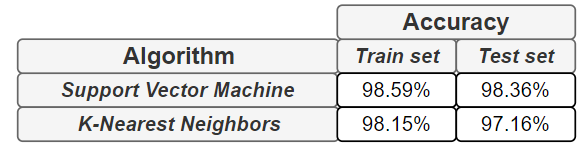}
    \caption{Accuracy on tabular data}
    \label{fig:baseline}
\end{figure*}
 Similarly, the same pre-processing phase was carried out. $\\$
The tabular data were splitted in train, test and evaluation set with a proportion of 30\% for test set. In addition, a percentage of 30\% of the test set was used for evaluation. 
The results obtained (shown in figure \ref{fig:baseline}) were higher than GCN and GAT applied on graph structures but are lesser than GraphSage results. Supporting the evidence, the test performances of the classification on tabular data are also associated to the Sensitivity Analysis results in the previous figures. 

\subsection{Ablation Study}



To better test the performances of our approach, we performed an ablatio study \cite{sheikholeslami2021autoablation,meyes2019ablation} considering  GraphSage, as it emerged as the best-performing model in the initial analysis phase and using the graph wich resulted in best performances, i.e. cosine similarity with a threshold of 0.95.

\begin{figure}[h]
    \centering
 \includegraphics[width=0.50\textwidth,keepaspectratio]{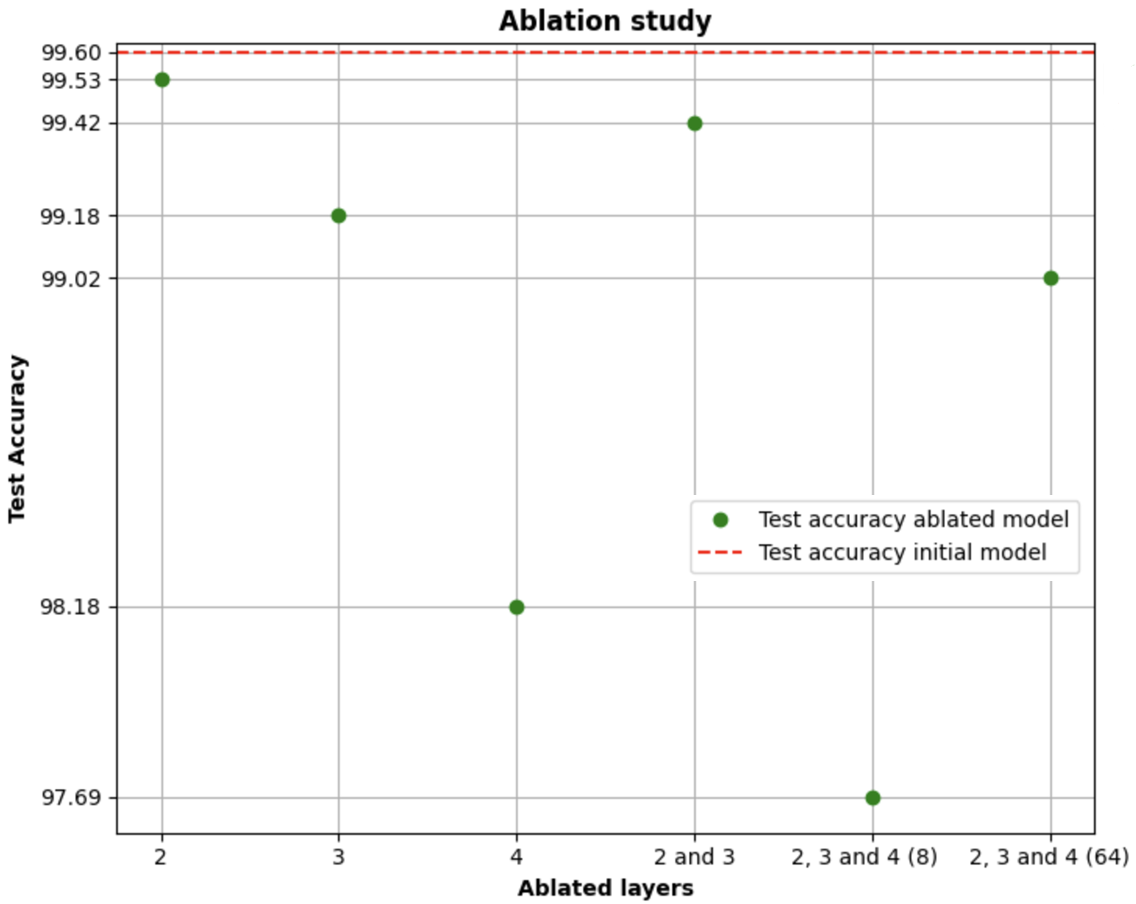}
    \caption{Results of ablation study}
    \label{fig:ablation}
\end{figure}

In this case, the study was conducted by removing one layer at a time, the second, third, and fourth layers, and finally all three layers, while varying the number of neurons and considering initially 8 neurons and then 64 neurons. In Figure \ref{fig:ablation} the results obtained are reported, on which it is possible to make the following considerations:
\begin{itemize}
\item the removal of  the fourth layer slightly reduces performance compared to removing the other layers.
\item the removal of the third layer reduces performance slightly more than removing the second layer.
\item the removal of the second, third, and fourth layers, considering 8 neurons, leads to the minimum accuracy value achieved in testing.
\item the removal of the second, third, and fourth layers, considering 64 neurons, reduces the accuracy value achieved in testing, but it remains higher than that obtained under the same conditions with 8 neurons and higher than that obtained by removing only the second or third layer.
\end{itemize}

Moreover, it could be argued that the fourth layer may be more important than others in aiding the task of node classification.

\section{Conclusion}

In this study, we introduced an innovative approach for managing patient triage in emergency departments through the application of artificial intelligence (AI) and network science. Utilizing machine learning algorithms and graph neural networks, our team crafted an AI module adapt at precisely categorizing patients into various triage levels. This system exhibited superior performance over conventional triage methods in our tests.

The findings from our research underscore the advantages of AI integration in healthcare, especially in the context of patient triage. This technological integration not only streamlines resource allocation but also minimizes errors in triage assessments. Our AI-driven method, which analyzes a patient's detailed medical history alongside their current vital statistics, enables a more detailed and accurate evaluation of their immediate medical needs. This approach significantly enhances the prioritization process of patients within emergency departments.

Moreover, our innovative strategy of representing patient data as graph nodes and employing graph neural networks for classification marks a notable advancement in the field of medical informatics. This technique not only boosts the precision of patient triage but also paves the way for novel methods of analyzing patient data in emergency healthcare scenarios.

To sum up, our study illuminates the transformative potential of merging AI with conventional healthcare methodologies to enhance both patient care and the operational efficiency of emergency departments. The adoption of AI-based systems in healthcare has the promise to redefine triage processes, ensuring more effective and optimized patient treatment. Looking ahead, the focus could be on refining these AI models and examining their applicability in diverse healthcare environments, which could further substantiate and expand upon the observed benefits of this study.

\section{Data and Code Availability}

All the data used in this article and the code for reproducing the experiments are available at \url{https://github.com/hguzzi/iatriage.git}. 


 \section{Acknowledgements}
 This work was funded by the Next Generation EU - Italian NRRP, Mission
4, Component 2, Investment 1.5, call for the creation and strengthening of
'Innovation Ecosystems', building 'Territorial R\&D Leaders' (Directorial
Decree n. 2021/3277) - project Tech4You - Technologies for climate change
adaptation and quality of life improvement, n. ECS0000009. This work
reflects only the authors' views and opinions, neither the Ministry for
University and Research nor the European Commission can be considered
responsible for them.

\end{document}